# Using a Convolutional Neural Network Model to Assess Paintings' Creativity

Zhehan Zhang,[1] Meihua Qian[1], Li Luo[2], Qianyi Gao[3], Xianyong Wang[4], Ripon Saha[5], Xinxin Song[6]

[1]College of Education, Clemson University
[2]School of Computing, Clemson University
[3]College of Education, University of Iowa
[4]College of Business, University of Clemson
[5]Department of Computer Engineering, Arizona State University
[6]Dipartimento di Architettura (DiDA), University of Florence

**Author Note**
Correspondence concerning this article should be addressed to Qianyi Gao, College of Education, University of Iowa. Email: qianyi-gao@uiowa.edu.

**Abstract**

The assessment of artistic creativity has been a long-standing challenge spanning decades of research. The traditionally human-rated tests of creativity such as the Torrance Figural Tests of Creativity (TTCT) and the Test of Creative Thinking -Drawing Production(TCT-DP) produce results that are reliable and valid to a relatively high degree. However, a key limitation of these traditional methods is that the scoring tends to be exceptionally time-consuming as it must be done manually by trained examiners. Therefore, many researchers have been trying to use the machine learning techniques to assess creativity, especially artistic creativity. For example, Cropley and Marrone (2022) applied a convolutional neural network (CNN) to assess students' responses to a figural creativity test - the TCT-DP, and Patterson et al. (2023) created an automated drawing assessment platform to extract visual creativity scores from simple sketches productions. However, existing studies focused on using machine learning models to assess the creativity of drawings which are based on figural creativity tests and are not typically considered paintings. Hence, this study aims to develop a CNN model capable of automatically evaluating the creativity of paintings. The training dataset consists of 80% paintings and the test dataset is the remaining 20%, with a total of 600 paintings created by professional artists and children. The model achieved around 90% accuracy and a significantly faster speed compared to human raters. We believe that the application of machine learning will open a new path forward for artistic creativity assessment.

# Using a CNN Model to Assess Paintings' Creativity

## Introduction

Creativity assessment has a long and rich history, despite the belief that some people think creativity is difficult to measure. Dating back to early 20th century's explorations of the human mind, the psychologists used the psychometric approaches to evaluate creativity, which in this period dominated the field. This approach focuses on four main areas: the creative process itself, personality traits and behaviors associated with creativity, the characteristics of creative outputs, and the environments that foster creativity (Plucker, Makel, & Qian, 2019). In the mid-20$^{th}$ century, J.P. Guilford (1950) addressed the American Psychological Association (APA) to focus on the then-neglected study of creativity. Guilford's call to action was instrumental in inspiring subsequent scholars to develop various methods for assessing creativity. Based on the work of Amabile and other researchers during the 1980s, who extended their methodologies to the study of creativity, the field has moved beyond simply examining each area in isolation. This was represented through a combination of findings and comprehensive theoretical framework to guide the field of educational administration. Nowadays, the field of creativity assessment is more active and dynamic than it has ever been before (Plucker et al., 2019).

One important subset of creativity assessment is artistic creativity assessment. It has been done in many ways. The artistic creativity of a person is the outcome of the complex relationship between the specific creative person, process, product, and environment. As Csikszentmihalyi mentioned in his book *The Systems Model of Creativity*,"Artistic creativity is as much a social and cultural phenomenon as it is an intrapsychic one" (Csikszentmihalyi, 2014, p. 228). Early studies of artistic creativity assessment (1950s-1970s) focused on evaluating artistic creativity based primarily on expert ratings of an art piece's novelty or unconventionality (Amabile, 1982). More recent work has looked at both the person, the cognitive processes involved, and the environment that enables creativity (Csikszentmihalyi, 2014). For decades, the assessment of artistic creativity has primarily relied on divergent thinking (DT) tests. The process of seeking out new solutions for a problem even if a person doesn't see any connection between its parts is called divergent thinking and this is one of the key aspects found in creative persons (Torrance, 1962). Although DT tests can evaluate the ability to generate ideas. Yet, this approach implicitly undervalues other aspects of creativity (Plucker et al., 2019).

Modern perspectives in the field of neuroscience challenge this idea, arguing that artistic creativity should be assessed comprehensively using a systems model distinct from scientific creativity. A voxel-based morphometry study revealed that the divergent thinking (DT) task, traditionally considered a crucial measure of artistic creativity, engages the Left Middle Frontal Gyrus (MFG) region of the brain. Interestingly, this finding suggests that the brain regions activated during DT tasks, which were previously thought to be linked to artistic creativity, are more akin to those brain regions involved in scientific creativity (Shi et al., 2017). Consequently, past approaches that heavily relied on expert ratings of divergent thinking (DT) products or any criteria related to DT may have significant limitations in accurately assessing artistic creativity. There is an urge that new measurement tools should be involved in the creativity assessment field, such as machine learning or AI.

**Definitions**

*Differences between Paintings and Sketches*

The past research focused on drawings, which are basically simple, single-colored sketchings, when it comes to assessing visual arts creativity. Traditional methods such as the Torrance Figural Tests of Creativity (TTCT) (Torrance, 1988) and the Test of Creative

Thinking-Drawing Production (TCT-DP) (Urban & Jellen, 1996; Urban, 2005) have been widely used. These human-rated tests produce reliable and valid results to a relatively high degree. Nevertheless, they are time-consuming and expensive to administer (Cropley & Marrone, 2022). Notably, our study is specifically designed for assessing the creativity of paintings created by professional artists and children. Comparing both datasets, paintings in our dataset (see the comparison in Figure 1) are apparently more complicated than simple drawings.

Kenyon Cox, a renowned American painter from the 19th century, defined painting as "the art of representing on a plane surface (in contrast to sculpture which works in three dimensions) the forms and colors of objects" (Cox, 1916). This definition brings out the general agreement on the nature of painting as an ancient art form that has been involved with applying shades, different colors to create a visually appealing composition.

**Figure 1**

*Comparison of Sketches in a typical creativity test and Paintings in our dataset*

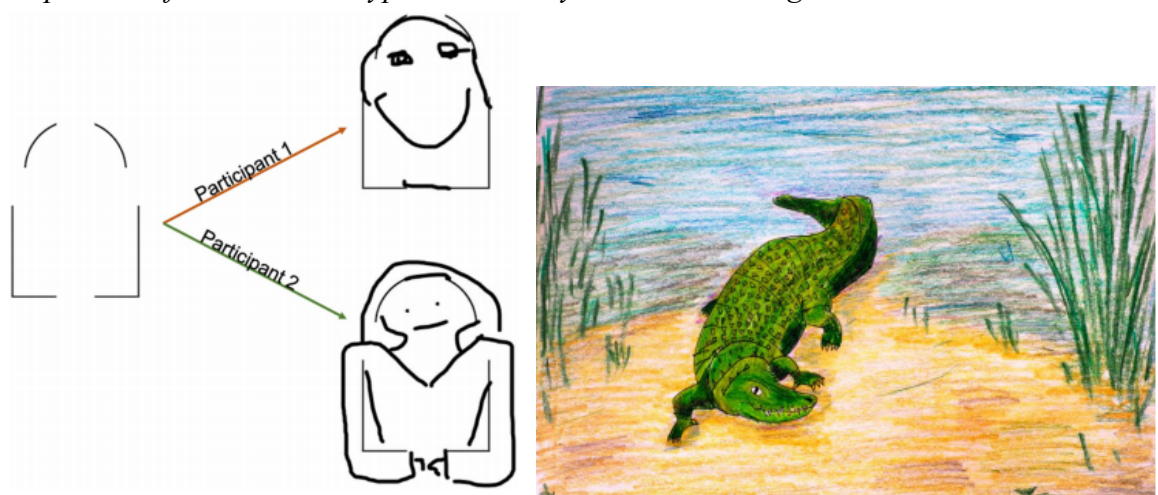

*Note.* The left picture is an example of Multi-Trial Creative Ideation (MTCI) Incomplete Shapes Task item and responses, which has two typical response sketches of figural creativity tests (Patterson et al., 2023). The picture on the right is an anonymous child's painting that received a creativity score of 55 out of 100, as evaluated by experts in our data samples.

***Assessment of Creativity in Paintings***

The assessment of creativity in paintings is the main part within the realm of artistic creativity evaluation. However, it has been largely neglected for decades. Historically, studies on artistic creativity have primarily focused on the broader concept rather than specifically addressing the assessment of paintings' creativity which is understandable. However, the methods used to evaluate the creativity of artworks may also be applicable to the assessment of paintings' creativity.

One of the additional problems that need to be considered before identifying the acceptance of artworks by the public is how people view creativity in artworks. Creativity is often seen as an inherently personal and context-dependent trait, which makes uniform assessments challenging (Alabbasi et al, 2022). Past researchers frequently use the Consensual Assessment Technique (CAT; Amabile, 1982). This approach is to ask human raters (ordinary people with proper training or creativity assessment experts) to provide their personal scores for each task following certain rubrics. Each rater's score reflects their unique perspectives and biases, which can vary widely, thereby accomplishing the achievement of a reliable consensus usually requires thousands of raters. Notwithstanding, it is very time-consuming and in need of huge human resources to support the research. Thus, there is a claim for a more efficient and cheap way to assess the creativity of paintings.

*Artificial Intelligence (AI) Tools, Machine Learning and Deep Learning*

Artificial intelligence (AI) is the study of constructing computer systems that behave intelligently and assist people to solve practical problems. Earlier attempts in AI were primarily theoretical, without many feasible real-world applications. In recent decades, AI has seen a surge of change with the combination of powerful machines and advanced machine learning methods (Jordan et al,2015). Within the AI field, machine learning (ML) related applications involving computer vision, natural language processing, speech recognition, and robotic controls have developed rapidly. Conventional machine learning methods rely on specific data mining algorithms, such as linear regression, logistic regression, decision trees, support vector machines, k-nearest neighbors, and naive Bayes. These ML methods have demonstrated effectiveness across various domains and tasks (Ertel,2011). However, as the exponential growth of data, ML has reached a boundary in its scalability and capability. Deep learning (DL), a subset of machine learning, is inspired by the processing patterns of human brains (Alzubaidi et al., 2021). Multiple neurons in deep neural networks can learn data iteratively and construct the optimal model for estimating new data. DL outperforms traditional ML in its ability to automatically learn complex representations. The relationship between these AI tools is illustrated in Figure 2.

**Figure 2**
*The Application of AI to Paintings' Creativity Assessment*

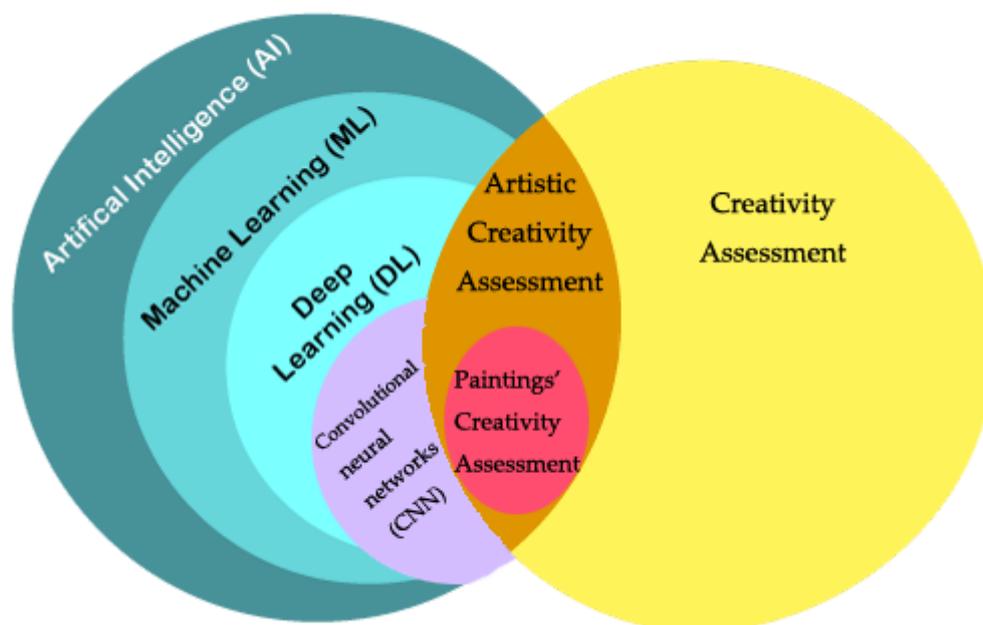

*Convolutional Neural Networks (CNN) models for image rating*

Convolutional Neural Network (CNN) is a specialized type of deep learning model designed specifically for processing structured grid data like images (Krizhevsky et al., 2012). Convolutional neural networks (CNNs) have been applied to image analysis and grading in various domains. Early studies have developed CNNs to evaluate the aesthetic quality and interestingness of generic images. In the most basic sense, Convolutional Neural Network is a variant of transfer learning, which is a part of machine learning. It has been also known as deep learning and its popularity is rapidly increasing because of its universal approach that enables it to do the jobs of other machine learning approaches and has the highest accuracy. At the beginning, the technology of convolutional neural networks was presented in the 1980s (LeCun & Bengio, 1995) but the practical use only happened recently with the introduction of rapid high-power processors as the primary hardware source. CNNs are used extensively for a variety of image processing applications, including object recognition (e.g., detecting traffic lights in pictures), face recognition, image enhancement and image effects (e.g., creating slow-motion effects in a video) (Cropley & Marrone, 2022).

We have found out that the Convolutional neural networks (CNNs) have revolutionized computer vision in recent years. By automating feature extraction, CNNs have significantly outperformed traditional methods in tasks like image classification, object detection, and semantic segmentation. It is one of the most popular tools in the AI field.

The existing information shows that some researchers use a CNN model to assess the creativity of simple sketchings. For instance, Cropley and Marrone (2022) applied image classification, not continuous scores, to focus on Divergent Thinking (DT) figural tests, which the tests provide prompts-incomplete shapes. Patterson et al. (2023) generated continuous scores; however, still focused on figural creativity tests and provided drawing prompts. Also they used a CNN model that is no longer state-of-the-art. These CNN models could reach around 90% accuracy compared to well-trained human raters. Therefore, it is a relatively high chance that we can use a similar CNN model to assess the creativity of professional artists and children's paintings which are more complicated drawings, with a high degree of accuracy and efficiency. And so far, a comparably extensively vetted solution for the assessment of paintings' creativity is very rare.

In summary, the integration of convolutional neural networks (CNNs) for automated image analysis represents a significant advancement in assessing creative outputs. This technology leverages deep learning algorithms to evaluate and rate images with a level of precision that approaches human judgment. As these systems continue to evolve, harnessing large datasets and computational power, they tend to expand the boundaries of how we understand and quantify visual creativity and aesthetics. This progress in the field of computer vision and machine learning enhances the efficiency of creativity assessments, and we believe that the application of machine learning will open a new path forward for artistic creativity assessment.

**Research Question**

The existing literature has largely focused on figural creativity within structured drawing tasks, which often lack complexity and diversity. The creativity of paintings, a more sophisticated and more intricate form of visual art, has been overlooked. Addressing this gap, our study investigates whether a CNN model can effectively assess the creativity of paintings produced by professional artists and children. Specifically, we ask this research question: **Can a CNN model be trained to evaluate creativity in paintings, and how closely do its assessments align with human judgments?** This research aims to develop automated creativity assessment methods and explore their application to complex artworks.

**Materials and Methods**

**Paintings**

The dataset consists of 600 paintings, with 400 paintings being created by children and purchased from Istock, and the rest being created by professional artists and downloaded from Wikimedia Common's website. The children's paintings are relatively poor in skills and similar in subjects (e.g., family, city life, and Astronauts in spaceships). However, the professional artists' paintings are well-depicted in terms of subjects and abundant in detail with various choices of subjects. We ask two experts who have the arts background to score from 1-100 points to assess the creativity of these 600 paintings following certain rubrics.

**Sample Selection**

This study involved analyzing a dataset comprising 600 paintings, divided into two distinct categories. The first category includes 400 children's paintings, which were acquired from iStock. These selections were made using the platform's embedded search engine, specifically filtering for the 'Kids Painting' category which features approximately 800 paintings. These artworks, originating from children worldwide, are accessible to anyone who registers an account on iStock and agrees to the platform's pricing terms. Due to budget constraints, only 400 paintings were purchased by selecting every second painting displayed in the search results. The quality of these paintings varies significantly; some exhibit advanced skills akin to those of professional artists, while most are noticeably less proficient. The subjects depicted in these paintings are diverse, reflecting the broad spectrum of children's imaginations, with common themes including family life, urban environments, astronauts, and playgrounds. All acquired paintings meet high standards of image clarity and definition.

The second category consists of 200 paintings by unknown artists, sourced from Wikimedia Commons. The selection process mirrored that used for iStock, utilizing an embedded search engine to filter results. The criteria for selection included a minimum resolution of 600x600 pixels to ensure the clarity of each painting, as blurred images could hinder detailed assessment by human raters. The first 200 paintings that met these criteria were included in the study.

This methodology ensures a robust and varied collection of paintings, facilitating a comprehensive analysis while adhering to the constraints of the study's budget and quality requirements.

**Ethical Considerations**

We have purchased 400 paintings from the website iStock for the purpose of training a machine learning model for non-commercial academic research. Permission for this use was confirmed through email communication with iStock staff. Additionally, we have sourced 200 paintings from Wikimedia Commons that are over 70 years old and whose authors are unknown. Based on U.S. copyright law-*Copyright law of the United States of America* (17 U.S. Code § 302 - Duration of copyright) and Wikimedia Commons guidelines (Wikimedia Foundation. (n.d.). Terms of use), these paintings are also approved for use in non-profit, academic research projects.

**Rubric**

To assess the creativity of a painting, normally we define artistic creativity as *originality+aesthetic factors* (Amabile,1979), these traditional evaluations often rely on human judgment to gauge artistic quality, considering elements like novelty, technique, and emotional impact. However, with the integration of machine learning tools into art analysis, we must adapt these subjective evaluations into measurable features that computational models can process. Drawing from the work of researchers like Lu et al. (2015), we apply computational thinking to assess creativity. Specifically, we break down the aesthetic factors of a painting into four key elements: Color, Texture, Composition, and Content. And with originality together, we developed a formula to assist human raters in assigning a score to each painting based on these components.

The evaluation of a painting's creativity is structured around five primary components: originality, color, texture, composition, and content. Each component is assessed on a scale from 0 to 20 points, contributing to an overall creativity score that ranges from 0

(lowest creativity) to 100 (highest creativity), with scores allocated as follows:

>Excellent (16-20 points)
>Good (11-15 points)
>Fair (6-10 points)
>Poor (1-5 points)

1. Originality (0-20 points)
>Originality involves the capacity to produce work that is both novel and appropriate (Amabile,1982). According to Ernst Gombrich (1966), originality in art is not merely about novelty; it concerns the ability to introduce new ideas, techniques, or styles that have a meaningful impact on the field of art. This includes the introduction of innovative concepts, the demonstration of high skill levels, and the creation of works that do not simply imitate others but offer new visual and thematic insights (Gombrich, 1966).

2. Color (0-20 points)
>Color evaluation focuses on lightness, colorfulness, color harmony, and distribution. The effective use of color should evoke emotional responses and captivate the viewer's attention, potentially altering their perception. Thoughtful selection and application of colors to create specific moods or atmospheres are crucial, reflecting the artist's intent and the painting's thematic elements.

3. Composition (0-20 points)
>The arrangement of elements within a painting, such as lines, shapes, colors, and textures, is critical. The composition should be harmonious and visually appealing, balancing the visual weight of the elements to create a stable and aesthetically pleasing piece. As noted by Gombrich (1995) in Art and Illusion, the composition's effectiveness is integral to the overall impact of the artwork (Gombrich, 1995).

4. Texture (0-20 points)
>Texture involves the use of materials and techniques to enhance the painting's depth, volume, and interest. Innovative approaches, unique brushwork, or unusual combinations of media can significantly add to the creativity of a painting. The texture should contribute to the realism or enhance the thematic expression of the artwork.

5. Content (0-20 points)
>Content considers the historical and cultural context of the artwork. Creative works may challenge societal norms, address contemporary issues, or reinterpret traditional themes in novel ways. The relevance of the subject matter to the painting's theme, its ability to convey a story or emotion, and its stylistic appropriateness are all evaluated under this criterion.

**Scoring**
>It is challenging to create a universal rubric for assessing all paintings because artistic styles and cultural norms evolve over time. Just like we would not use the same criteria to evaluate works by Picasso and Michelangelo, whose approaches were shaped by different artistic movements and cultural contexts, it would be inappropriate to apply a single rubric to all genres art paintings., as Csikszentmihalyi challenged the notion of a "timeless, constitutional artistic personality," suggesting instead that the traits considered artistically creative are largely influenced by the prevailing styles and cultural dynamics of the time (Csikszentmihalyi, 2014). Therefore, the rubric is mainly applicable to classical art paintings, not suitable for some controversial modern art paintings. The application of these criteria is exemplified through the assessment of various types of paintings, ranging from children's artworks to pieces by ordinary and famous artists. For instance, a children's painting of a dog might score lower overall due to simplistic execution and lack of depth in originality and content (See Figure 3 and 4). In contrast, a well-executed piece from the Baroque period might score higher due to sophisticated use of color, texture, and composition, though it may receive a lower score in originality if it adheres closely to the typical style of its era without introducing new elements.

**Figure 3**
*Example of Low Creativity Painting from a Child*

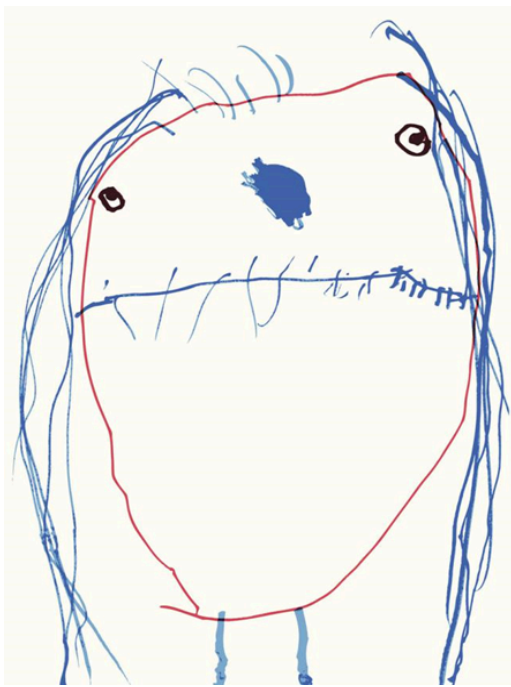

*Note.* The painting was originally from Istock.com, with permission to use in this study.

**Figure 4**
*Example of High Creativity Painting from a Professional Artist*

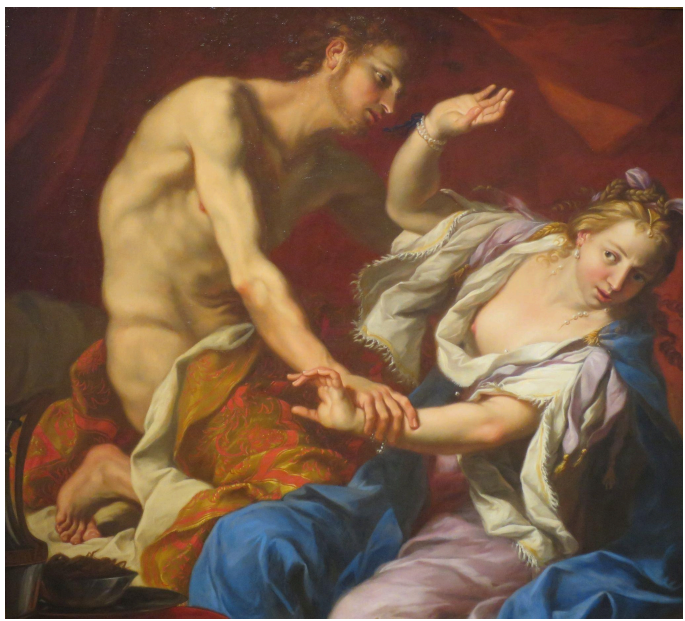

*Note.* The painting was originally from Wikimedia Commons, complying with local laws and regulations to use in this study.

**Human Rating**

The evaluation of the 600 paintings was conducted by two doctoral students, serving as independent raters. The first rater is a doctoral candidate at Clemson University, who holds an MA in Film and Drama Studies from one of China's most prestigious normal universities. The second rater is a doctoral student at the University of Florence who specializes in Art Design and has also earned an MA in the same field from the same institution.

Prior to commencing their evaluations, both raters were thoroughly briefed on the scoring rubrics designed for this study. The intraclass Correlation Coefficient between two human raters for our 600 paintings dataset was 0.99.

**Model Training**

The Efficientnet CNN model was used to assess the creativity of the paintings. There are various CNN architectures like AlexNet, VGGNet, GoogLeNet, ResNet and more, each building upon previous models and achieving higher accuracy and efficiency on benchmark databases (Canziani et al., 2017). In recent years, researchers seek to focus on parameter efficiency. For individual researchers, EfficientNet has better accuracy with fewer parameters and fast speed for training. In our paper, we choose the B1 version of EfficientNet (Tan et al., 2019) as a deep learning model which is generalizable as the baseline model of EfficientNet, also it is smaller and faster than ResNet-152. EfficientNet is mainly composed of multiple basic convolutional operations called MBConv (Sandler et al., 2018; Tan et al., 2019). MBConv stands for MobileNetV2 block convolution, which is lightweight CNN architecture but further improves efficiency. The architecture sequentially consists of one convolutional layer, seven MBConv blocks, followed by convolutional and pooling layers and a fully connected layer. As input images go through multiple convolutional layers, initial layers extract from low level feature maps while deeper layers extract from higher level feature maps. Low level features are like edges and textures, while higher level features represent abstract and complex visual concepts. Through a multi-iteration training process, the model can learn well from the image features.

It's noticeable that the CNN for image classification is widely utilized in multiple domains beyond the computer vision field. However, as we investigated, the CNN model for creativity domain in the education field is a novel trial (Cropley & Marrone, 2022; Patterson et al., 2023). We focus on the research of painting's creativity assessment among children and artists, specifically inputting the paintings, learning visual features from multi-convolutional layers, rating creativity scores by using numerical regression models. The existing EfficientNet model is well trained for ImageNet classification. To adjust the model for our research, there are two problems we need to solve. For one, the datasets we use are paintings and pictures which are different from ImageNet. The technique is to utilize fine-tuning in transfer learning that transfers the knowledge from the source dataset to the target dataset (Zhang et al., 2020). Assuming the parameters of pre-trained EfficientNet contain knowledge of the ImageNet dataset, it is also applicable to our painting's datasets. On the other hand, in order to assess paintings, we apply linear regression to map the output features of multi-convolutional layers to numerical scores. We created a new model based on the pre-trained EfficientNet and replaced the output classification layer with a linear regression layer. Besides, due to the input paintings data having various sizes and ratios of width and height, we transform these paintings by cropping the central pixels based on minimal value of both width and height and resizing them into 720 x 720 pixels square. Then for expanding the training and testing datasets, we do image augmentation by randomly horizontal and vertical flipping for the input data. This data preprocessing effectively improves the deep learning process and results.

The experiment works on the Palmetto platform of Clemson University using V100 and 16GB GPU. For the training part, since our rubrics of artistic creativity rate as five criterions, that is, originality, color, texture, composition and content, our model outputs the scores among these five dimensions and optimizes them to minimum through training iterations. The loss function we choose is mean square loss, which measures the average squared difference between the predicted score and human rates' score.

We divided the datasets into 80% training data and 20% testing data, which is a commonly used ratio by practitioners (V. Roshan Joseph, 2022). For testing data, we sample them by the region spanned which is better than random testing split. Specifically, out of a total of 400 children's paintings and 200 artists' paintings, we allocated 320 children's paintings and 160 artists' paintings for training, while the remaining paintings were reserved for testing. Our approach involved selecting every 5th painting for testing. As the total size of the dataset is relatively small, we determine not to use a validation set for hyperparameters tuning. Hyperparameters cannot learn from training but set to work critically in the training process. Through experiment, we set mini-batch size to 10 images and learning rate to 0.0005 best works for our model.

**Figure 5**
*Our CNN model architecture, based on the EfficientNet Architecture*

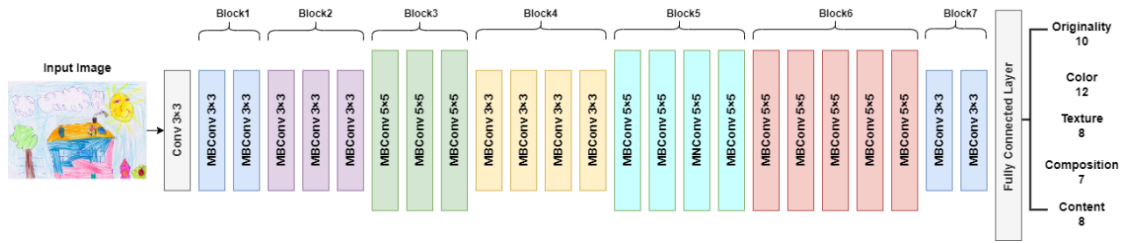

Evaluation metrics: The accuracy and efficiency of the CNN model were evaluated using a variety of metrics for regression prediction, including the mean absolute percentage error, correlation and r-squares. The CNN model shows strong correlation with human raters achieving 0.956 with 95% confidence interval [0.94, 0.97], r-square value of 0.912 and mean absolute percentage error (MAPE) of 14.95%. Those metrics indicate our CNN model fits very well with human raters. Our results involved dividing the test data into five classification models, each assessed by percentage accuracy and confusion matrix. We utilize a combination of five rubric components to assess the creativity of paintings.

## Results

**Correlation between human ratings and the CNN model-generated scores**

In a total of 600 paintings, we considered the held-out strategy for 120 paintings (20% of dataset) as the testset and evaluated the CNN model based on the test paintings. The CNN model-generated scores were pretty close to the human ratings'. Figure 6 shows the model's strong correlation with human raters (correlation coefficient of 0.956) further supports its potential to emulate human judgment accurately.

**Figure 6**
*Our Model Prediction with Human Ratings' Correlation*

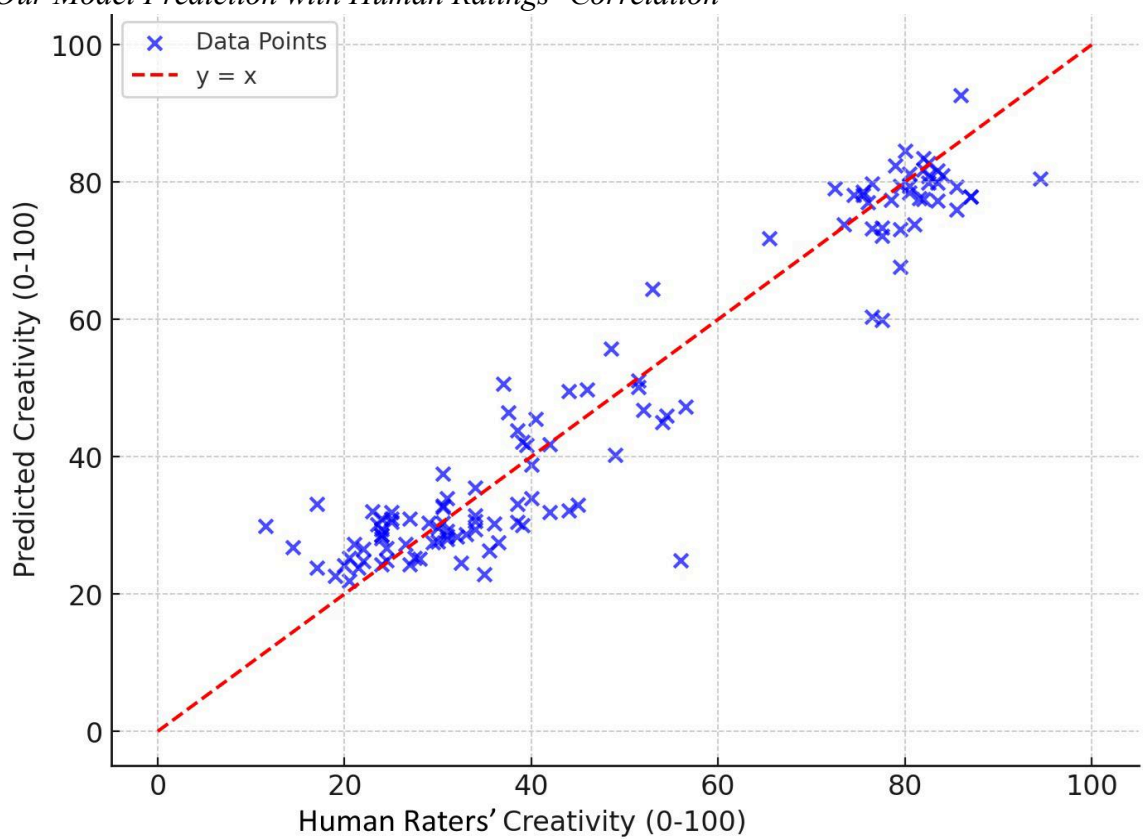

**Models' Percentage Accuracy**

Here, we divided the test dataset into different models for detailed comparison. The accuracy of a model for each class is measured by the proportion of test images correctly assigned to different creativity classes. One important aspect of model training is the number of epochs, which refers to how many times the model processes the entire training dataset. An epoch represents a complete pass through the entire training dataset during model training. For instance, 100 epochs mean the model has processed the dataset 100 times. Generally, more epochs lead to improved model accuracy, while increasing the number of epochs can improve accuracy, gains tend to plateau after around 200 epochs for the models in this study. Another crucial hyperparameter is the learning rate. It controls the size of the model's parameter updates in response to the estimated error after each epoch. It also influences both the speed and precision of the model's learning process. A higher learning rate accelerates training but risks instability, while a lower rate ensures stability but might slow down training. The default learning rate for our models was 0.0005, as we have explained in the previous model training part. In the following accuracy tables, red numbers mean the amount of paintings were mis-classified by our CNN model.

*Model 1: Low and High Creativity*

Model 1 was conducted by splitting the paintings dataset into two subsets, categorized as low and high creativity. Based on our rubric, the intervals for low and high creativity are divided into two approximately equal ranges (0-57) and (58-100). Totally 382 and 218 paintings were labeled as low and high creativity, respectively. The actual class represented the average scores of human raters while the predicted class represented the CNN model-generated scores. Model 1 achieved a high accuracy of 99.17% and only mis-classified one painting out of all 120 paintings, indicating excellent performance of the CNN model in classifying paintings into low and high creativity categories.

Table 1
*Model 1 Accuracy*

| Actual class | Predicted class | |
|---|---|---|
|  | Low | High |
| Low | 79 | 1 |
| High | 0 | 40 |

*Model 2: Low, Medium, and High Creativity*

Model 2 increased the complexity by introducing a medium category for creativity assessment, dividing into three intervals: low (0-35), medium (36-57), and high (58-100). We classified 224 paintings as low creativity, 158 as medium creativity, and 218 as high creativity in all 600 paintings.

Table 2
*Model 2 Classes*

| Class | Score Range | n |
|---|---|---|
| Low | <35 | 224 |
| Medium | 36-57 | 158 |
| High | ≥58 | 218 |

Out of 120 paintings' test dataset, the model achieved an accuracy of 91.67%. The confusion matrix (see Table 3) indicates a robust ability to distinguish between low and high creativity scores, with a minor challenge in distinguishing between low and medium categories, as evidenced by 10 misclassifications. The model's precision in classifying the medium category was admirably high, with few errors leading to an effective capture of nuanced distinctions in creativity. This model suggested that adding a medium category may help capture more subtle distinctions in creativity scoring, although at the cost of a slight decrease in overall accuracy.

Table 3
*Model 2 Accuracy*

| Actual class | Predicted class | | |
|---|---|---|---|
|  | Low | Medium | High |
| Low | 51 | 0 | 1 |
| Medium | 10 | 18 | 0 |
| High | 0 | 0 | 40 |

### Model 3: Low to High Creativity

In Model 3, the dataset was further refined by introducing four categories of creativity: low (0-35), medium (36-57), medium high (58-71), and high (72-100), with respective samples of 224, 158, 28, and 190 paintings.

Table 4
*Model 3 Classes*

| Class | Score Range | n |
|---|---|---|
| Low | $\leq 35$ | 224 |
| Medium | 36-57 | 158 |
| Medium High | 58-71 | 28 |
| High | $\geq 72$ | 190 |

The model reported an accuracy of 88.33% through the test dataset. The confusion matrix revealed excellent performance in identifying high creativity, but showed some confusion in the medium ranges with a few misclassifications between adjacent categories. Specifically, the model struggled slightly with distinguishing between low and medium categories and between medium and medium high categories. This model suggested that while further segmentation of the creativity spectrum allows for a more nuanced understanding, it also introduced more complexity, which may challenge the model's predictive capabilities, as indicated by the reduced accuracy compared to Models 1 and 2.

Table 5
*Model 3 Accuracy*

| | Predicted class | | | |
|---|---|---|---|---|
| Actual class | Low | Medium | Medium High | High |
| Low | 51 | 0 | 1 | 0 |
| Medium | 10 | 18 | 0 | 0 |
| Medium high | 0 | 0 | 1 | 3 |
| High | 0 | 0 | 0 | 36 |

### Model 4: Very Low to Very High Creativity

Table 6
*Model 4 Classes*

| Class | Score Range | n |
|---|---|---|
| Very Low | $\leq 15$ | 6 |
| Low | 16-35 | 218 |
| Medium | 36-57 | 158 |
| High | 58-71 | 28 |
| Very High | $\geq 72$ | 190 |

Model 4 expanded the categorization to five groups: very low (0-15), low (16-35), medium (36-57), high (58-71), and very high (72-100). It classified smaller subsets, including a very low category consisting of only 6 paintings. The model achieved an accuracy of 86.67%. The confusion matrix shows that while the model was highly accurate in identifying the high creativity class, it had minor difficulties with the very low and low categories, possibly due to the small sample size in the very low category. This expanded model indicates that increasing the number of categories to include a 'very low' class does not necessarily enhance performance significantly, as the accuracy slightly decreased due to the finer distinctions required by the model.

Table 7
*Model 4 Accuracy*

| Actual class | Predicted class | | | | |
|---|---|---|---|---|---|
| | Very Low | Low | Medium | Medium High | High |
| Very low | 0 | 2 | 0 | 0 | 0 |
| Low | 0 | 49 | 0 | 0 | 0 |
| Medium | 0 | 10 | 18 | 0 | 0 |
| Medium High | 0 | 0 | 0 | 1 | 3 |
| High | 0 | 0 | 0 | 0 | 36 |

Model 4 Results Summary:
Accuracy 86.67%
Interval [0,16) [16,36) [36,58) [58,72) [72,100]
Number of each class: 6, 218, 158, 28, 190

**Model 5: Six Levels**
Table 8
*Model 5 Classes*

| Class | Score Range | n |
|---|---|---|
| Very Low | <16 | 6 |
| Low | 16-36 | 218 |
| Medium | 36-58 | 158 |
| Medium High | 58-72 | 28 |
| High | 72-90 | 185 |
| Very High | ≥90 | 5 |

Model 5 further detailed the classification by introducing a six level category, very high (90-100), for a total of six classes: very low (0-16), low (16-36), medium (36-58), medium high (58-72), high (72-90), and very high (90-100). This model processed the smallest number of paintings in the new category, with only 5 classified as very high. It achieved an overall accuracy of 85%. The performance was robust in classifying the high and very high categories but showed some limitations in differentiating the lowest categories, likely due to the limited number of examples in extreme categories. The introduction of a very high category highlights the model's ability to recognize exceptional creativity, though at the expense of a slight reduction in overall accuracy due to the greater complexity and finer distinctions in creativity assessment.

Table 9
*Model 5 Accuracy*

| Actual class | Predicted class | | | | | |
|---|---|---|---|---|---|---|
| | Very Low | Low | Medium | Medium High | High | Very high |
| Very low | 0 | 2 | 0 | 0 | 0 | 0 |
| Low | 0 | 49 | 0 | 0 | 0 | 0 |
| Medium | 0 | 10 | 18 | 0 | 0 | 0 |
| Medium High | 0 | 0 | 0 | 1 | 3 | 0 |
| High | 0 | 0 | 0 | 0 | 34 | 1 |
| Very high | 0 | 0 | 0 | 0 | 1 | 0 |

Model 5 Results Summary:
Accuracy 85%
Interval [0,16) [16,36) [36,58) [58,72) [72,90) [90,100]
Number of each class: 6, 218, 158, 28, 185, 5

In conclusion, our series of classification models effectively demonstrated the capability of neural networks to assess the creativity of paintings. Starting with a simple binary classification in Model 1, which achieved high accuracy, then we gradually increased the complexity by introducing more categories in subsequent models. While this allowed for a more detailed analysis of creativity, it also led to a slight reduction in accuracy. Models 2 and 3 struck a good balance between detail and performance, maintaining over 88% accuracy. Models 4 and 5, with even more categories, faced challenges due to the finer distinctions and smaller sample sizes in extreme categories, resulting in slightly lower accuracy.

**Discussion**

The results obtained from this study suggest that the CNN model employed can effectively automate the assessment of creativity in paintings, achieving an average accuracy of 90.17% in these five models. By leveraging the capabilities of deep learning algorithms, specifically CNN models, it becomes possible to develop advanced tools and systems that can objectively evaluate the creativity and artistic merits of paintings. By developing a structured rubric that includes originality, color, texture, composition, and content, our study introduced a comprehensive framework for evaluating artistic creativity. The model's strong correlation with human raters (correlation coefficient of 0.956) further supports its potential to emulate human judgment accurately.

The findings of this study align with and build upon previous research in automated creativity assessment using convolutional neural networks (CNN). Cropley and Marrone (2022) employed a CNN to assess responses to figural creativity tests, specifically focusing on divergent thinking through the analysis of incomplete shapes in simple drawings. Similarly, Patterson et al. (2023) developed an automated drawing assessment platform to evaluate creativity in basic figural productions. However, those studies were limited by the oversimplification of their data-stock sketches and incomplete drawings which did not capture the complexity of the real professional artwork. On the one hand, our study is a further development of the CNN models because it applies them to more complex and varied datasets including both children's and professional paintings. This wider perspective makes it possible to see artistic creativity in a more complex way, like texture, color, and composition. Our model achieved near accuracy results but the augmented capability of dealing more complex creative levels is its main distinction. This point at CNN models, if tuned well, can go beyond the limitations of previous studies that were outdated and only focused on simple forms of creativity. Consequently, the present study greatly enhances the use of machine learning in creativity evaluation by focusing on the evaluation of paintings, which has been an underexplored area in previous literature.

Our work makes significant contributions to the field of creativity assessment, particularly through the innovative application of CNNs to evaluate artistic creativity in paintings. This pioneering approach opens new avenues for leveraging deep learning in creativity research, addressing the critical limitation of traditional methods being time-consuming and resource-intensive. The development of a detailed and structured rubric provides a standardized framework that can be used in educational and research settings, promoting more objective and replicable assessments.The potential applications of such tools are vast and far-reaching. Art educators could employ these tools to provide objective feedback and assessments to students, helping them develop their creative skills and artistic techniques. Additionally, art galleries, museums, and other cultural institutions could benefit from these tools by automating the evaluation process for identifying and curating creative and innovative artworks (Winner & Martino, 2000).

By bridging the gap between computer science and creativity research, our study exemplifies the interdisciplinary potential of modern AI technologies, setting a strong foundation for future research and encouraging further exploration into more complex models and diverse datasets.

**Limitation and Future Study**

This study has demonstrated the potential of a CNN model for automating the assessment of creativity in paintings. By leveraging the power of deep learning algorithms, researchers have taken a significant step towards developing advanced tools that can automatically evaluate artistic creativity. Future research should explore the use of various kinds of CNN models for other creative tasks, such as assessment of famous paintings.

However, it is crucial to acknowledge the limitations of the current study. The relatively small size of the dataset, consisting of only 600 paintings, and the composition of the dataset itself may introduce biases or limitations in the generalizability of the findings. One major limitation is the absence of creativity score samples within the 63-78 range due to lack of funding purchasing highly skilled children's paintings. Therefore, it is essential to include more children's paintings with high creativity scores to further validate and strengthen the conclusions drawn from this study. Future research should focus on utilizing larger and more diverse datasets that encompass a broader range of artistic styles, genres, and cultural contexts.

By expanding the dataset and incorporating a wider variety of artistic expressions, researchers can ensure that the CNN model's capabilities are thoroughly tested and refined, resulting in a more robust and reliable tool for assessing creativity in paintings. In the meantime, the CNN model itself achieves high accuracy on the training dataset compared to the test dataset, which suggests a potential overfitting issue to generalize unseen data. This indicates the model might be too tailored to the training data and less effective to predict new paintings. And future studies could explore the potential integration of additional data sources, such as artist biographies, historical contexts, and art criticism, to enhance the model's understanding and evaluation of creativity.

Furthermore, the findings of this study open up an exciting avenue for future exploration in the realm of creativity assessment tasks beyond the assessment of paintings. Researchers could investigate the potential applications of CNN models in evaluating creativity across various artistic domains, such as music, sculpture, and design. By continually pushing the boundaries of what is possible with machine learning algorithms, we can unlock new maps in our understanding of human creativity.